\documentclass[letterpaper]{article} 
\usepackage{aaai24}  
\usepackage{times}  
\usepackage{helvet}  
\usepackage{courier}  
\usepackage[hyphens]{url}  
\usepackage{graphicx} 
\usepackage{natbib}  
\usepackage{caption} 
\usepackage{algorithm}
\usepackage{algorithmic}

\usepackage{xcolor} 
\usepackage{amsthm} 
\usepackage{amsfonts} 
\usepackage{amsmath} 
\usepackage{subfig} 
\usepackage{comment} 

\usepackage{amssymb} 

\usepackage{newfloat}
\usepackage{listings}
\DeclareCaptionStyle{ruled}{labelfont=normalfont,labelsep=colon,strut=off} 
\floatstyle{ruled}
\newfloat{listing}{tb}{lst}{}
\floatname{listing}{Listing}
\title{Learning Performance Maximizing Ensembles with Explainability Guarantees}
\author{
    Vincent Pisztora, 
    Jia Li
}
\affiliations{
    Department of Statistics, Pennsylvania State University, USA\\

    uxp5@psu.edu, jol2@psu.edu
}

\usepackage{bibentry}

\begin{document}

\newtheorem{lemma}{Lemma}
\newtheorem{proposition}{Proposition}
\newtheorem{theorem}{Theorem}
\newtheorem{definition}{Definition}

\maketitle

\begin{abstract}
In this paper we propose a method for the optimal allocation of observations between an intrinsically explainable glass box model and a black box model. An optimal allocation being defined as one which, for any given explainability level (i.e. the proportion of observations for which the explainable model is the prediction function), maximizes the performance of the ensemble on the underlying task, and maximizes performance of the explainable model on the observations allocated to it, subject to the maximal ensemble performance condition. The proposed method is shown to produce such explainability optimal allocations on a benchmark suite of tabular datasets across a variety of explainable and black box model types. These learned allocations are found to consistently maintain ensemble performance at very high explainability levels (explaining $74\%$ of observations on average), and in some cases even outperform both the component explainable and black box models while improving explainability.
\end{abstract}

\section{Introduction}

In most high stakes settings, such as medical diagnosis \cite{gulum2021review} and criminal justice \cite{rudin2019stop}, model predictions have two viability requirements. Firstly, they must exceed a given global performance threshold, thus ensuring an adequate understanding of the underlying process. Secondly, model predictions must be explainable. 

Explainability, however defined \cite{linardatos2020explainable}, is a desirable characteristic in any prediction function. Intrinsically interpretable ``glass box" models (\cite{agarwal2021neural}, \cite{pmlr-v130-lemhadri21a}, \cite{rymarczyk2020protopshare}), which are explainable by construction, are particularly advantageous as they require no additional post-hoc processing (\cite{ribeiro2016lime}, \cite{lundberg2017unified}) to achieve explainability, and thus also avoid complications arising from post-hoc explanation learning (\cite{rudin2019stop}, \cite{garreau2020explaining}). Due to these advantages, glass box models are uniquely suited to settings where faithful explanations of predictions are required.

However, using an approach of ``complete explainability", in which a glass box model is used as the prediction function across the entire feature space, may not be viable. It may be the case that, in a given setting, no glass box exists that can adequately model the relationship of interest in all regions of the feature space. Thus in some regions, the model's predictions will fail to exceed the performance threshold required by the use-case. If, as a consequence, the model exceeds the application's global error tolerance (e.g. a low accuracy in stroke prediction \cite{gage2001validation}), it may not be usable in practice.

\begin{figure}%
    \centering
    \includegraphics[width=8.42cm]{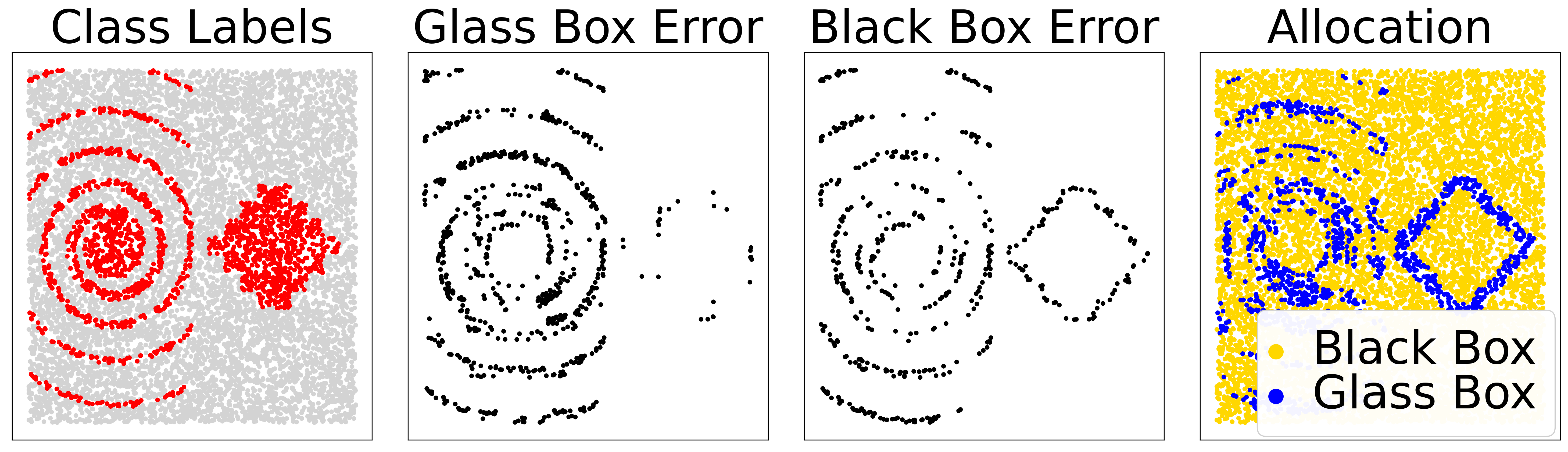}
    \caption{This figure shows a two-class classification task in which the areas of expertise (the diamond pattern for the glass box and the spiral pattern for the black box model) are complementary. The glass box achieves a $92.7\%$ accuracy, the black box reaches $95.0\%$ accuracy, and the allocated ensemble of the two exceeds both with a $95.8\%$ accuracy. Thus, the resulting EEG allocation improves performance over both component models while also providing explainability (for $20\%$ of observations in this case).}%
    \label{fig:toy}%
\end{figure}

An alternative, ``partial explainability" approach requires instead that only a proportion of observations be provided intrinsically explainable predictions. We will refer to this proportion, which is the proportion of observations for which the explainable model is the prediction function, as the explainability level $q$. Such approaches, including our proposed method, Ensembles with Explainability Guarantees (EEG), can provide high performance while maximizing explainability, and work especially well in cases where the explainable model can be paired with an alternate model with complementary strengths. As demonstrated in Fig. \ref{fig:toy}, by identifying the areas of expertise of the glass box and black box models, the EEG approach can allocate predictions accordingly to improve both performance and explainability.

\begin{figure*}[ht]
    \centering
    \includegraphics[trim={1cm 3cm .75cm 3cm},clip,scale=0.4]{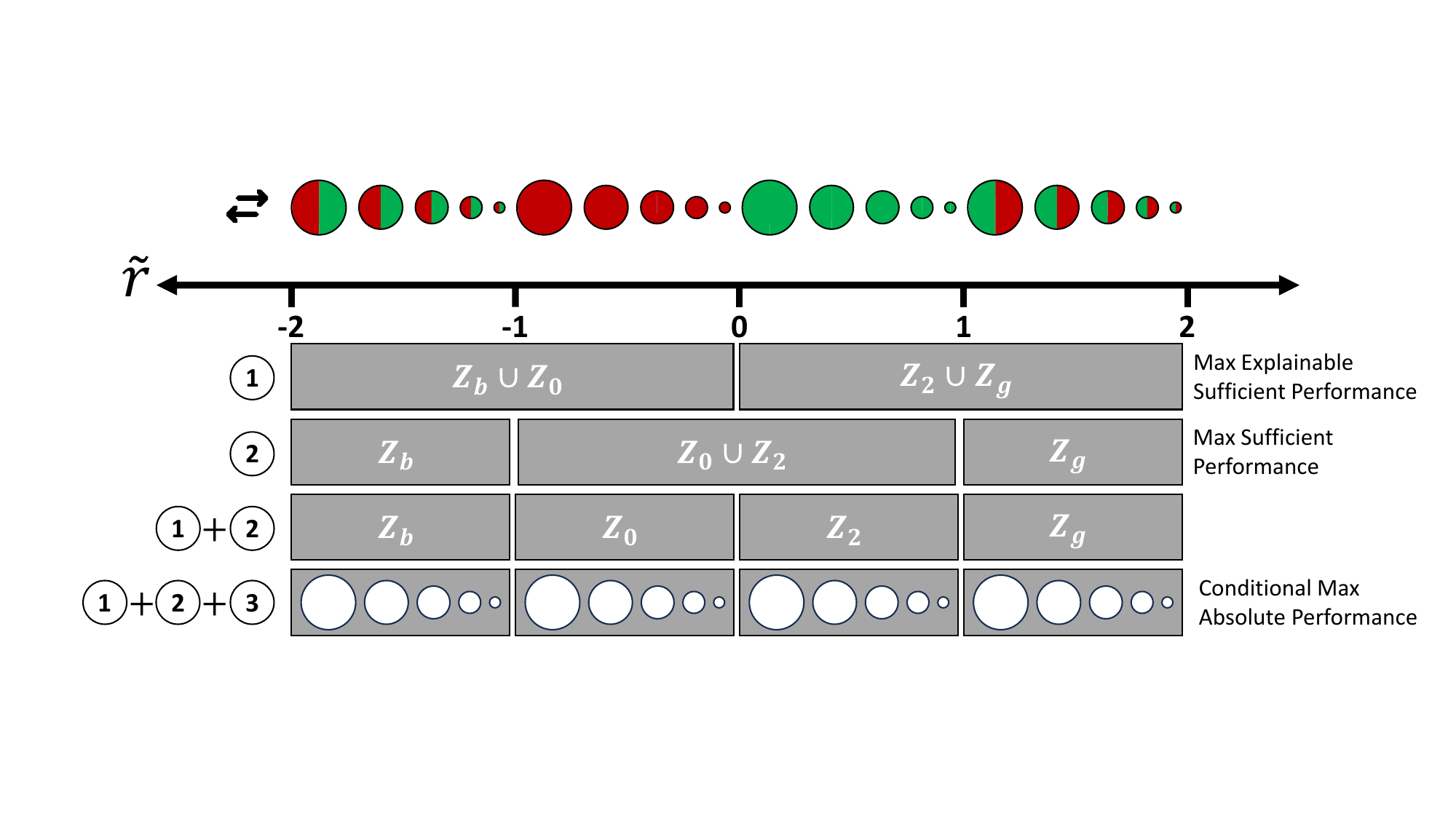}
    \caption{This figure provides an intuition for how the proposed allocator $a^{\prime}_q(v)$ ranks observations. This ranking can be seen as ordering the sufficiency sets ($Z_b, Z_0, Z_2, Z_g$) step-by-step, with each step achieving a particular optimality condition (formalized in Propositions \ref{thm_1_perfmax}, \ref{thm_2_perfmaxg}, and \ref{thm_3_absperf}). In the figure, individual observations are circles above the number line with size representing the magnitude of the relative advantage of black box over glass box, the left-hand color representing sufficiency of the glass box, and the right-hand color representing the sufficiency of the black box (green for sufficient, red for insufficient).}
    \label{fig:introeeg2}
\end{figure*}

Generally, implementations of a partial explainability approach consist of an ensemble of models including at least one explainable model and alternate model (often a black box model), and an allocation scheme by which observations are distributed among the ensemble members for prediction. Individual methods are characterized by their heterogeneity in the following aspects. 

Methods vary in the range of component models they can accommodate. Some are defined for only one set of glass box, black box, and allocator model types - for example LSP \cite{wang2012localspacepart} and OTSAM \cite{wang2015treeadd}, which use binary tree-type splitting to define regions, and linear models and sparse additive models respectively to predict within regions. Other methods are black box agnostic but still limited in glass box and allocator model type - for example HyRS \cite{wang2019gainfree}, HyPM \cite{wang2021hybrid}, CRL \cite{pan2020bbcompan}, and HybridCORELS \cite{ferry2023learning}, which use rule-based models as both glass box and allocator. EEG is the only fully model-agnostic partial explainability method which can be implemented with any combination of glass box, black box, and allocator models.

Methods also vary in the approach used to learn each ensemble member model (i.e. glass box and black box). Most methods first learn the black box model globally (on the full dataset), and then learn the glass box model locally (on its allocated subset of the data), either simultaneously with the allocator (HyRS, HyPM, CRL, and HybridCORELS) or in an alternating EM-style (LSP, OTSAM, and AdaBudg \cite{nan2017adaptive}). EEG on the other hand, learns both ensemble member models globally first before learning the allocations between them - similar to most general adaptive ensembling methods, e.g. \cite{gao2019adaptensb_2}, \cite{inoue2019adaptensb}. 

Finally, methods are characterized by their allocation criteria which commonly consist of an objective which combines one or more of the following - the explainability level, the underlying task performance of the ensemble, and the complexity of the glass box model. Most methods optimize a measure of post-allocation ensemble performance - LSP, HyRS, HyPM, and HybridCORELS minimize a 0/1 misclassification loss, AdaBudg uses a more flexible logistic loss, and CRL maximizes accuracy across a range of explainability levels. Several methods with rule-based glass box/allocator hybrid models (HyRS, HyPM, CRL, and HybridCORELS) also include a penalty on the complexity of these models. To control the explainability level, methods either include a reward term in the loss (HyRS, HyPM, and CRL), or directly restrict the model space to candidates which achieve the explainability level (HybridCORELS). In contrast, EEG optimizes an MSE loss between the predicted and actual ``glass box allocation desirability" percentile of each observation.

More extensive reviews of the partial explainability approach and explainability methods in general are available in \cite{linardatos2020explainable}, \cite{nauta2022anecdotal}, and \cite{9551946}.

As outlined above, our proposed method, Ensembles with Explainability Guarantees (EEG), differs from existing works in its approach to the partial explainability problem. The key novelties of this new approach, and their corresponding advantages are summarized below.

\textbf{Independent and Global Component Models:} The first key innovation of the EEG approach is the independent learning of each component model (i.e. the ensemble member models and allocator). As a result, EEG is agnostic to task, data, and component model type. Thus, the most powerful models can be used for each component as determined by the setting - in contrast with previous works which are more restricted. 

Another important consequence of separate component model learning is that glass box predictions are independently explainable in the global context, and thus immune from ``explainability collapse" - a scenario in which the allocator subsumes the glass box's prediction role, diminishing the value of the explainable prediction, in the extreme case reducing the glass box to an uninformative constant function. On the other hand, methods which either learn glass box models locally, or jointly with the allocator, are vulnerable to this type of degeneration. 

\textbf{Allocation Desirability Ranking:} The second novel aspect of the EEG approach is the concept of allocation desirability. Given an ensemble of models, allocation desirability quantifies how beneficial it is for a given observation to be allocated to the default ensemble member model, say the glass box. Thus, it induces a preference for glass box allocation between all pairs of observations and consequently also defines a ranking of allocation preference across all observations that is optimal irrespective of the desired explainability level. 

A key advantage of such a ranking is that it is independent of the training criteria of the ensemble member models, and thus can be adapted to score allocation desirability using metrics that best fit the setting. Indeed, the EEG desirability metric builds a ranking using a combination of relative sufficient performance and absolute performance measures which can natively accommodate any underlying problem type (e.g. regression, classification). This particular desirability metric also offers several additional benefits including allocation desirability percentile and sufficiency category estimates for each observation. 

\textbf{Q-Complete Allocation Optimality:} The final key point of novelty of the EEG approach is the optimality of allocation, as defined in Propositions \ref{thm_1_perfmax}, \ref{thm_2_perfmaxg}, \ref{thm_3_absperf}, and \ref{thm_monotone}, which is encoded in the allocation desirability ranking for any explainability level. Thus, the learned allocator, which estimates this ranking, is an explicit function of $q$ and provides the allocation solution to any explainability level after training only once. This capability is in contrast with previous works which provide, at most, several explainability level solutions with varying degrees of stability \cite{ferry2023learning}.

These unique capabilities of the EEG method enable the following practical use cases:
\begin{itemize}
    \item Given a minimum performance requirement on the underlying task, the method can be used to obtain the allocation with the highest explainability level that achieves or exceeds the performance threshold.
    \item Given a minimum explainability level requirement, the method can be used to obtain the allocation with the highest ensemble performance which meets or exceeds the required explainability level.
    \item Given a minimal level of post-allocation glass box-specific performance, the allocation that achieves the highest explainability level while meeting or exceeding this requirement can be found.
    \item Given a set of observations, sufficiency category estimates can be obtained for each, identifying which observations are likely to yield incorrect decisions and describing the likely failure mode for each such case to inform potential post-hoc remedies.
\end{itemize}

In the following sections, we first describe our method in detail and provide some theoretical assurances on the characteristics of the resulting allocator in the Methods section. Then, in the Experiments section, we describe the experimental settings and the estimation of the allocator, and demonstrate the method's favorable performance.

\section{Methodology}

\subsection{Setting}

First, we define the underlying task as the estimation of the function $f(x) = y, f\in\mathcal{F}=\{f:\mathcal{X}\rightarrow \mathcal{Y} \}$ where $x\in\mathcal{X},y\in\mathcal{Y}$. We also define observations as $z=(x,y)\in \mathcal{X}\times\mathcal{Y}=\mathcal{Z}$, the training dataset $D^n=\{z_i:i\in\{1,...,n\},z_i\in\mathcal{Z}\}$, and loss function for the underlying task $l:\mathcal{Y}\times\mathcal{Y}\rightarrow \mathbb{R}$. Next, we define the ensemble component models - first, the intrinsically explainable glass box model as $g: \mathcal{X}\rightarrow\mathcal{Y}$ and the alternate, black box model as $b: \mathcal{X}\rightarrow\mathcal{Y}$, both of which are learned independently on the full training dataset $D^n$. 

Next, we define the allocation task. We define the class of all allocator functions as $A=\{a:\mathcal{V} \rightarrow \{0,1\}\}$ and the class of all ``proper" allocator functions as $\hat{A}=\{a:\hat{\mathcal{V}} \rightarrow \{0,1\}\}$, where $\mathcal{V}=\mathcal{Z}\times \mathcal{W}$ is a general space of inputs, $\hat{\mathcal{V}} \subseteq \mathcal{V}$ contains only information available at allocation time, and $V^n = \{v_i : i\in \{1,...,n\}, v_i \in \mathcal{V}\}$. Next we define the class of $q$-explainable allocators as $A_q = \{a_q:a_q\in A, \frac{1}{n}\sum^{n}_{i=1}a_{q}(v_i) = q\}$ and the corresponding class of ``proper" $q$-explainable allocators as $\hat{A}_q = \{a_q:a_q\in \hat{A}, \frac{1}{n}\sum^{n}_{i=1}a_{q}(v_i) = q\}$, for $q \in \mathcal{Q} = \{\frac{i}{n}:i\in\{1,...,n\}\} \subseteq [0,1]$, with $q$ being the explainability level. Note, the set $A$ is used to define the optimal allocator, whereas the set $\hat{A}$ is searched to obtain an estimator of this optimum.

We next define indicators of performance sufficiency. These functions $s:\mathcal{F}\times \mathcal{Z} \rightarrow \{0,1\}$ should be thought of as context-dependent indicators of whether performance within a region of the feature space is sufficiently high to use the model in question reliably for explanation. Although the EEG approach holds for any such function $s$, sufficiency functions used in the Experiments section are defined as follows. For classification tasks, we define performance sufficiency as $s^{}_{f}(z)=\mathbb{I}{\{f(x)=y\}}$, and for regression tasks as $s^{}_{f}(z)=\mathbb{I}{\{l(f(x),y)<\epsilon\}}$. In practice $\epsilon$ should be selected based on problem-specific context, however, lacking such context in the regression experiments conducted for this study, $\epsilon$ was selected to be the lower of the average validation losses of $g$ and $b$, as a reasonable threshold for prediction correctness. These sufficiency indicators generate the following partition of the data: 
$Z^{}_0 = \{z:z\in \mathcal{Z}, s^{}_g(z)+s^{}_b(z)=0 \}, 
Z^{}_2 = \{z:z\in \mathcal{Z}, s^{}_g(z)+s^{}_b(z)=2 \}, 
Z^{}_g = \{z:z\in \mathcal{Z}, s^{}_g(z)=1,s^{}_b(z)=0 \}$, 
and $Z^{}_b = \{z:z\in \mathcal{Z}, s^{}_g(z)=0,s^{}_b(z)=1 \}$, with
$n_0 = |Z^{}_0|, n_2 = |Z^{}_2|, n_g = |Z^{}_g|, n_b = |Z^{}_b|$, and $n_q = nq$.

Next we motivate the use of the sufficiency perspective. Sufficiency functions are critical for defining coherent allocations when, as is often the case, the absolute performance measures used to learn ensemble component models do not match allocation preference (e.g. loss minimization vs accuracy maximization). Consider the constrained allocation decision in Table \ref{table:suff_eg}, in which only one observation can be allocated to $g$.

\begin{table}[t] 
\centering
\begin{tabular}{ |c|c|c|c|c| }
 \hline
 Obs & $l(g)$ & $s_g$ & $l(b)$ & $s_b$\\
 \hline
  $z_1$ & 0 & 1 & 2 & 1  \\
 \hline
  $z_2$ & 3 & 1 & 4 & 0  \\
 \hline
\end{tabular}
\caption{This table describes the loss values ($l$) and sufficiencies ($s$) of two observations ($z_1$, $z_2$), for both a glass box ($g$) and black box model ($b$). In the constrained allocation case, in which only one observation can be allocated to $g$, the optimal allocation changes depending on whether loss or sufficiency is used to determine allocation preference.}
\label{table:suff_eg}
\end{table}

In this case, loss minimization dictates an allocation of $z_1$ to $g$ and $z_2$ to $b$, which would allocate $z_2$ to an insufficient prediction. Sufficiency maximization would however yield a more satisfactory allocation of $z_2$ to $g$ and $z_1$ to $b$. This example demonstrates the utility of sufficiency allocation - distinguishing between a case where the user is willing to sacrifice ``a bit of performance" (as quantified by sufficiency) for explanation ($z_1$), and a case where even a small performance drop results in an explanation that is not sufficiently trustworthy to use ($z_2$).

In the next section, we define the objective of the allocation task and introduce our proposed approach for addressing it.

\subsection{Optimal Allocation}

In the allocation task, the objective is to construct an allocator $a_q$ that will determine which model, either the explainable $g$ or the black box $b$, is used for prediction on any given observation $z$, in a manner that is optimal relative to the following criteria. Firstly, for any given explainability level $q$, the allocator should distribute observations in a way that maximizes sufficient ensemble performance, defined as
\begin{gather}\label{eq:suff_perf}
S(a_q) = \frac{1}{n}\sum^{n}_{i=1}s(a_q,v_i)\\
s(a_q,v) = s_g(z)a_q(v)+s_b(z)(1-a_q(v))
\end{gather}
Secondly, and again for any $q$, the allocator should maximize sufficient explainable prediction, defined as
\begin{gather}\label{eq:cond_suff_perf}
S_{g}(a_q) = \frac{1}{n}\sum^{n}_{i=1}s_g(z_i)a_q(v_i)
\end{gather}
i.e. the performance of the model $g$ on the subset of observations it has been allocated. Thirdly, for any $q$, the allocator should maximize the ensemble absolute performance (as measured by $L(a_q)$ below), conditional on achieving the first two sufficient performance objectives. 
\begin{gather}\label{eq:abs_perf}
\begin{split}
L(a_q&) = \frac{1}{n}\sum l_i(a_q)\\=\frac{1}{n}\sum l(g(x_i),y_i)a_q(&v_i)+l(b(x_i),y_i)(1-a_q(v_i))
\end{split}
\end{gather}
Finally, the allocator should also be consistent in its allocations across the values of $q$, meaning that if an observation is allocated to $g$ for a given $q$, it should remain allocated to the glass box for all higher explainability levels as well.
Next, we define our allocator, and show that it meets the criteria introduced above. 

Our proposed allocation function $a^{\prime}_q(v)$ is defined as follows. 
\begin{equation}\label{eq:alloc}
    a^{\prime}_q(v)=\mathbb{I}\{r^{}(z)>1-q\}
\end{equation}
With rescaled ranking $\tilde{r}$ defined as
\begin{equation}
    \tilde{r}(z)=\frac{\mathrm{rank}_{D^n}(r^{}(z))}{n}
\end{equation}
and ranking $r$ defined as
\begin{equation}
    \begin{split}
    r(z)=2s^{}_g(z)-s^{}_b(z)-\sigma(l(g(x),y)-l(b(x),y))\\
    \end{split}
\end{equation}
where $\sigma(x)=\frac{1}{1+e^{-x}}$.

The intuition behind the allocator is depicted in Fig. \ref{fig:introeeg2} and is as follows. First, all observations are sorted in sufficient performance maximizing order, i.e. allocation of observations in $Z_g^{}$ to $g$ is prioritized over allocation of observations in $Z_2^{}$ and $Z_0^{}$, which in turn are prioritized over $Z_b^{}$. Next, observations are sorted in explainable sufficient performance maximizing order, i.e. $Z_2^{}$ is prioritized ahead of $Z_0^{}$ for allocation to $g$. Then, within each sufficiency category, observations are ordered in absolute performance maximizing order, i.e. observations with large relative performance of $g$ over $b$ are prioritized for allocation to $g$. Next, this ranking is normalized, yielding the glass box allocation desirability percentile $r^{}$. An important feature of this percentile is that it is constant with respect to $q$, thus the optimal observations to allocate to $g$, for any level of $q$, are simply the $n_q$ most highly ranked, resulting in the allocator $a^{\prime}_q$.

Note that in the described methodology, sufficiency based allocation can be viewed as a generalization of allocation via absolute performance, and can thus be reduced to the latter by selecting, for example, either $s^{}_{f}(z)=0$ or $s^{}_{f}(z)=1$, $\forall z,f$. 

Next, we state the optimality properties of the proposed allocator $a^{\prime}_q$. The proofs are available in the Theoretical Results section of the Appendix.

\begin{proposition}{(Maximal Sufficient Performance)}\label{thm_1_perfmax}\\
$\forall q \in \mathcal{Q}, a^{\prime}_{q} \in A^{*}_{q}$ where $A^{*}_{q}=\{a^{*}_{q} \colon a^{*}_{q}=\arg\max_{a_{q}\in A_{q}} S(a_q)\}$
\end{proposition}

In words, Proposition \ref{thm_1_perfmax} states that across all explainability levels $q$, the proposed allocator $a^{\prime}_q(v)$ (Eq. \ref{eq:alloc}) achieves the highest possible sufficient performance (as defined in Eq. \ref{eq:suff_perf}).

\begin{proposition}{(Maximal Sufficient Explainable Performance)}\label{thm_2_perfmaxg}\\
$\forall q \in \mathcal{Q}, a^{\prime}_{q} \in A^{*}_{q\vert g}$ where $A^{*}_{q\vert g}=\{a^{*}_{q\vert g} \colon a^{*}_{q\vert g}=\arg\max_{a_{q}\in A_{q}} S_{g}(a_q)\}$
\end{proposition}

Proposition \ref{thm_2_perfmaxg} states that, again across all explainability levels, the proposed $a^{\prime}_q(v)$ achieves the highest possible sufficient performance on those observations which have been allocated to the glass box $g$ (i.e. maximizing Eq. \ref{eq:cond_suff_perf}).

\begin{proposition}{(Conditional Maximum Absolute Performance)}\label{thm_3_absperf}\\
$\forall q \in \mathcal{Q}, a^{\prime}_{q} \in A^{*}_{q\vert s}$ where $A^{*}_{q\vert s}=\{a^{*}_{q\vert s} \colon a^{*}_{q\vert s}=\arg\min_{a_{q}\in A^*_{q}\cap A^*_{q\vert g}} L(a_q)\}$
\end{proposition}

Proposition \ref{thm_3_absperf} states that across all explainability levels, the proposed allocator $a^{\prime}_q(v)$ achieves the highest absolute performance (in the sense of Eq. \ref{eq:abs_perf}) available among all allocators that meet the conditions of Prop \ref{thm_1_perfmax} and Prop \ref{thm_2_perfmaxg}. 

\begin{proposition}{(Monotone Allocation)}\label{thm_monotone}\\
$\forall q_i<q_j \in \mathcal{Q}$, $\{v:v\in\mathcal{V},a^{\prime}_{q_i}(v)=1\}\subseteq \{v:v\in\mathcal{V},a^{\prime}_{q_j}(v)=1\}$
\end{proposition}

Finally, Proposition \ref{thm_monotone} states that the allocator $a^{\prime}_q(v)$ is self-consistent in the sense that if an observation is considered best allocated to the glass box at a given explainability level $q_i$, then is also guaranteed to remain allocated to the glass box at any higher higher explainability level $q_j$. This prevents the unintuitive phenomenon of an observation "hopping" in and out of the glass-box allocated bucket at different explainability levels.

\section{Experiments}
In this section we describe the data, model training procedures, performance evaluation metrics, and results of our experiments.

\subsection{Datasets}
Tabular data is used to evaluate the proposed methodology as it the setting for which the required intrinsically explainable glass box models are most readily available.
Following the tabular data benchmarking framework proposed by \cite{grinsztajn2022tree}, we conduct experiments on a set of 31 datasets (13 classification, 18 regression). These datasets represent the full set of provided datasets with quantitative features less the four largest scale datasets (omitted due to computational limitations). These datasets are summarized in Section B of the Appendix. 

Each dataset is split (70\%, 9\%, 21\%) into training, validation, and test sets respectively, following \cite{grinsztajn2022tree}. All features and regression response variables are rescaled to the range [-1,1].\\

\subsection{Models}
Both glass box and black box models are learned on the full training dataset for each underlying task. For classification datasets, two types of glass box model are fitted, a logistic regression and a classification tree, as well as two types of black box model, a gradient boosting trees classifier and a neural network classifier. Analogously, for regression datasets, two types of glass box model are fitted, a linear regression and a regression tree, as well as two types of black box model, a gradient boosting trees regressor and a neural network regressor. In all cases, the architecture of the neural networks is the ``Wide ResNet-28" model \cite{zagoruyko2016wide} adapted to tabular data with the replacement of convolutional layers with fully connected layers. 

An allocator is subsequently also learned on the full training dataset. Both gradient boosting trees regressors and neural networks are fitted as allocators for each allocation task.\\ For allocator training, the features $x$ are augmented with four additional constructed features, the predictions $g(x)$ and $b(x)$, and two distance measures $d(g(x),b(x))$ between them, the cross-entropy and MSE. In our experiments, inclusion of these features improved allocator learning - likely by removing the need for the allocator to attempt to learn these quantities on its own. Allocation performance is further improved by ensembling the feature-dependent learned allocator $a^{\prime}_q$ with a strong feature-independent allocator $a^{\prime\prime}_q$, where $a^{\prime\prime}_q(d(g(z),b(z)))=\mathbb{I}\{\frac{\mathrm{rank}_{D^n}d(g(z),b(z))}{n}<q\}$. $a^{\prime\prime}_q$ can be viewed as an ``assume the black box is correct" allocation rule which is more likely to assign an observation to $g$ if the distance between the predictions of $g$ and $b$ is low. Which of the two allocators is used for a given $q$ is determined by their respective performances on the validation set.

\subsection{Hyperparameter Tuning}
Hyperparameter tuning for all models is done using 4-fold cross-validation, with the exception of the neural network tuning which is done using the validation set. A grid search is done to select the best hyperparameters for each model with search values available in Section C of the Appendix. \\
Each glass box and black box model is tuned on the full set of hyperparameters each time it is replicated. The gradient boosting trees allocator models are retuned on the full hyperparameter set each time as well. The neural network allocator is not retuned however, and instead uses the optimal settings found in the fitting of the black box on each dataset.

\subsection{Metrics}

We define the following metrics which are used to measure performance of our method. First we define the Percentage Performance Captured over Random (PPCR) for a given allocator as follows: $PPCR(a_q) = \frac{AUC(a_q)-AUC(r_q)}{AUC(o_q)-AUC(r_q)}$ where $AUC(f_q)$ is the area under the curve of function $f_q$ over all values of $q$ in its domain, $o_q$ is the oracle allocator which has perfect information on the whole dataset, and $r_q$ is the random allocator which selects a subset from the data being allocated uniformly at random. The PPCR metric is a percentage and represents the proportion of the oracle AUC, in excess of that also covered by random allocation, that the learned allocator is able to capture. Thus a value of zero indicates performance on par with $r_q$ and a value of one represents perfect allocation.

Next, we define the Percent Q Equal or Over Max (PQEOM) as the percentage of $q$ values for which the allocator is performing at least as well as the most accurate ensemble member model (i.e. $g$ or $b$) and Percent Q Over Max (PQOM) as the percentage of q values for which the allocator is performing better than the most accurate ensemble member model (i.e. $g$ or $b$).

Next, we define the Percent Contribution of Feature-dependent Allocator (PCFA) as the percentage of $q$ values for which the feature dependent allocator $a^{\prime}_q$ is used for allocation decisions as opposed to the feature-independent allocator $a^{\prime\prime}_q$. A value close to one indicates that $a^{\prime}_q$ is used often, while a value close to zero indicates it is $a^{\prime\prime}_q$ instead.

Next, we define the $95\%$ Threshold Q Max (95TQM) as the highest value of $q$ for which the ensemble performance meets or exceeds $95\%$ of the performance of the better of $g$ and $b$. Thus this is a measure of how much explainability can be utilized before the performance price becomes material. 

Next, we define the maximum accuracy achieved by the allocator across all $q$ (Max Acc), and the highest value of $q$ for which this accuracy is maintained (Argmax $q$). The Max Acc can be benchmarked against the AUC, interpretable as the average accuracy across $q$. Each of these metrics is a percentage and higher values correspond with higher performance and higher explainability at this maximum performance level, respectively.

Finally, we define the accuracy with which the four sufficiency categories ($Z^{}_g$, $Z^{}_b$, $Z^{}_2$, and $Z^{}_0$) can be estimated as the sufficiency accuracy ($s$ Acc). The higher this accuracy, the better able the allocator is to inform the user of which category a given observation is likely to be a member of.

\subsection{Results}

\begin{figure}%
    \centering
    \subfloat[\centering PolR Dataset]{{\includegraphics[width=3.75cm]{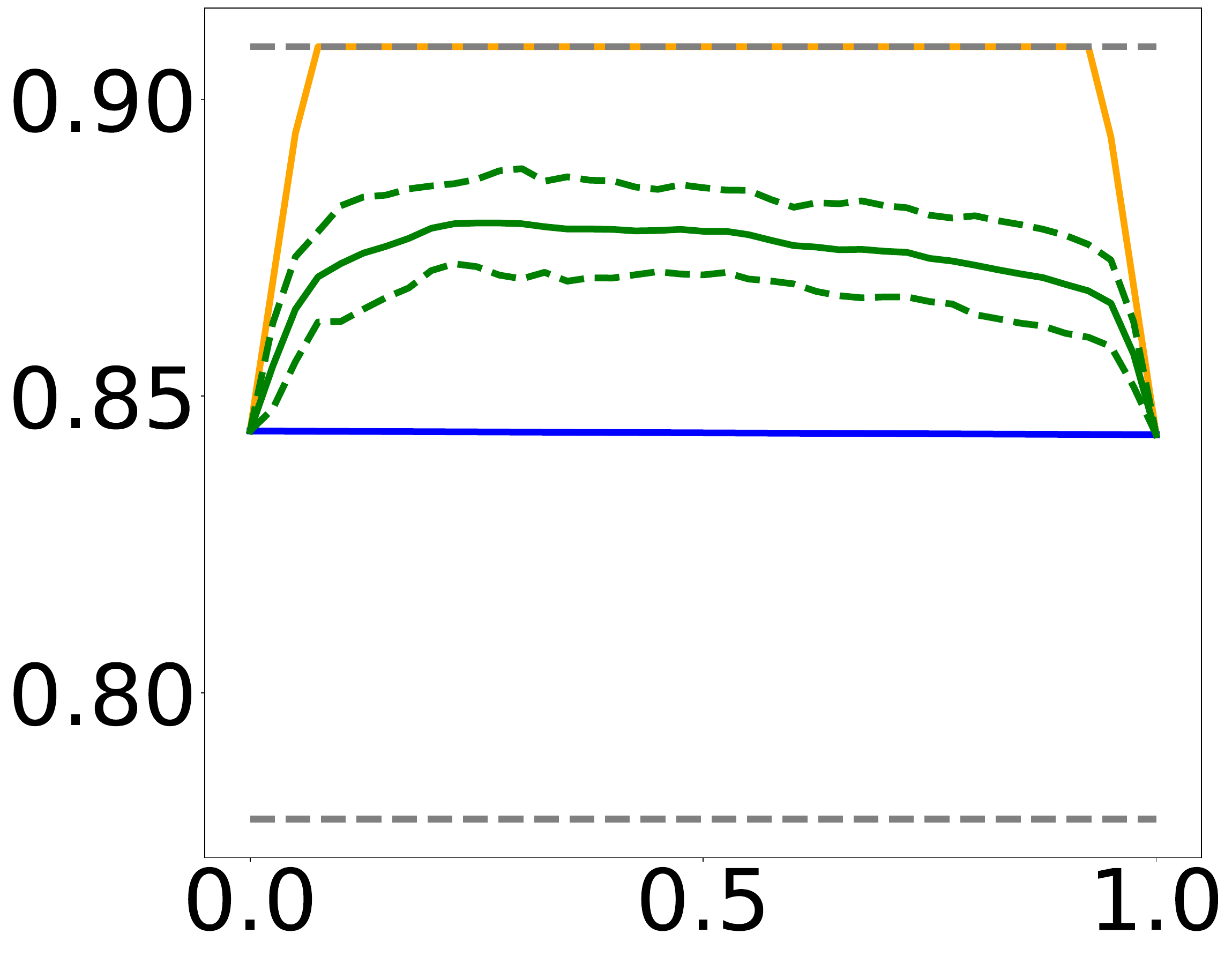} }}
    \qquad
    \subfloat[\centering SuperconductR Dataset]{{\includegraphics[width=3.75cm]{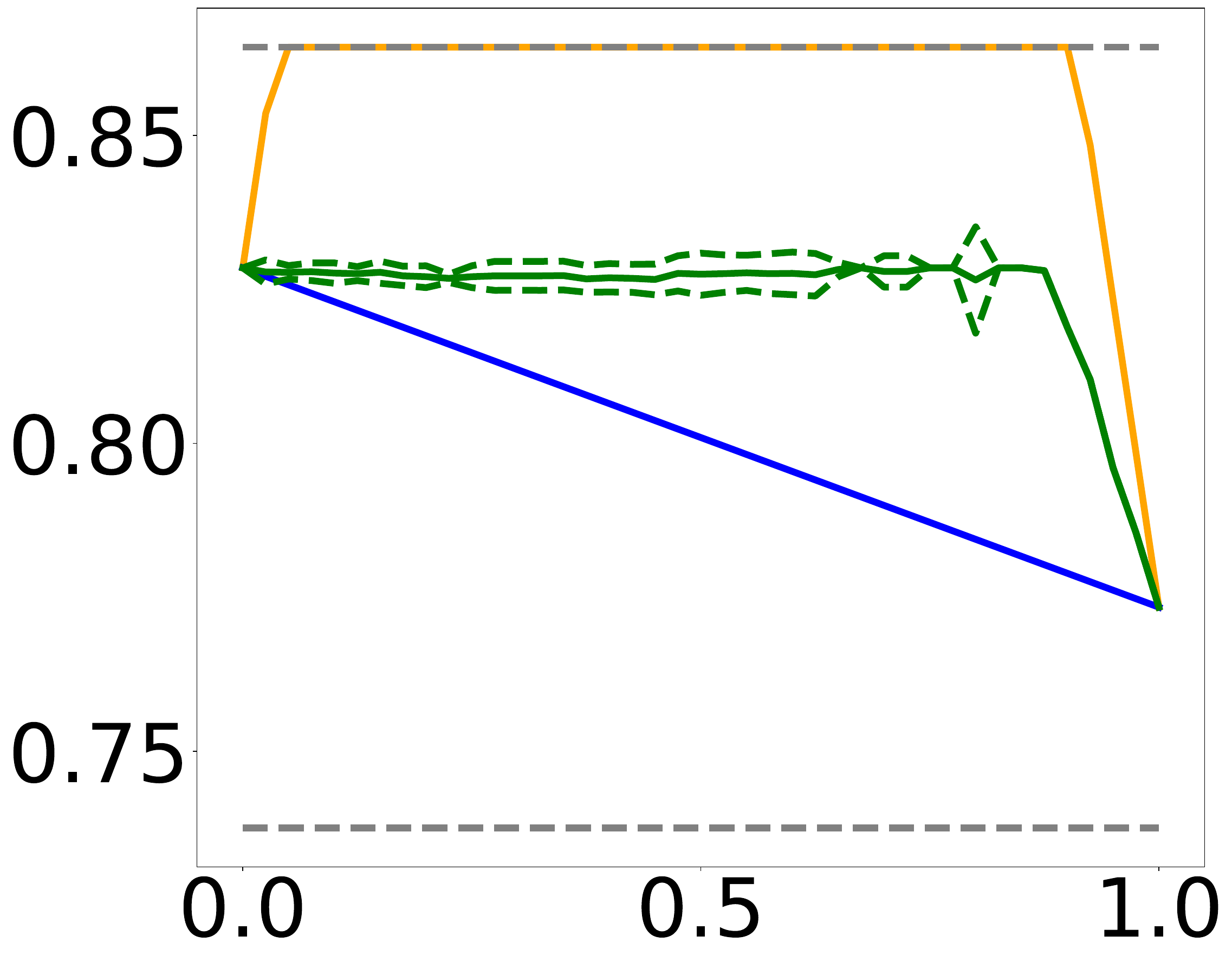} }}%
    \caption{This figure shows two examples of the explainability (x-axis) vs sufficient performance (y-axis) trade-off, comparing the random (blue), oracle (orange), and learned (green) allocation curves. The PolR dataset is an example of complementary $g$ and $b$ models, resulting in an allocated ensemble that outperforms both component models across most of the $q$ range. The SuperconductR dataset is an example of an explainability ``free lunch" in which the $b$ accuracy is maintained while increasing explainability using the allocator. Curves for all datasets are available in Section D of the Appendix}%
    \label{fig:curves}%
\end{figure}

Evaluation of allocator performance using the metrics defined previously as well as visual inspection of the performance vs explainability trade-off curves (Fig. \ref{fig:curves}) revealed both the benefits and some of the limitations of learned allocation in the tabular data setting. 

Firstly, performance was found to consistently and significantly outperform random allocation, as quantified by a cross-dataset PPCR of $37\%$ (Table \ref{table:results}), indicating that the learned allocation captured close to $40\%$ of the area under the curve available and in excess of random allocation. It was also found that on some datasets in particular, learned allocation performed close to oracle allocation (e.g. $89\%$ and $71\%$ on the IsoletR and BrazilianHousesR datasets). 

Learned allocation was also found to perform at least at the level of the best ensemble member model across an average of $74\%$ of the explainability range (PQEOM in Table \ref{table:results}). This indicates that for many datasets, there is a substantial explainability ``free lunch" to be taken advantage of without performance loss. On a few datasets, performance of the allocated ensemble was found to outperform both $g$ and $b$ for a majority ($93\%$) of the $q$ range (PolR and FifaR PQOM). The 95TQM metric also supported these conclusions, with a cross-dataset average value of $94\%$ indicating that allocation performance was within $5\%$ of maximal individual model performance across approximately all values of $q$.

Assessing the PCFA metric suggests some limits to the upside of learned, feature dependent allocation - at least in the tested tabular data setting. A cross-dataset average value of $35\%\pm34\%$ indicates that on average, the range for which the feature dependent allocator is used over the feature-independent one is indistinguishable from zero. This is consistent with a visual inspection of the representative performance-explainability curve e.g. Fig. \ref{fig:curves} (b) where there is no improvement to be had in excess of $a^{\prime\prime}_q$. However, it is noted that the only possibility for ``homerun" allocations is through the feature dependent $a^{\prime}_q$ as seen in Fig \ref{fig:curves} (a) with the PolR dataset and also in Table \ref{table:results} for datasets SulfurR, BikeSharingR, and FifaR. Thus the ensembled allocation scheme offers this upside without downside risk of low performance in either $a^{\prime}_q$ or $a^{\prime\prime}_q$.

Evaluation of the case in which a single allocation is needed is also positive. On a cross-dataset average, the $84\%$ maximal accuracy achieved is quite high, and is also achieved at a high average explainability level ($64\%$). Particularly strong individual results can be seen in the Pol and SulfurR datasets (Table \ref{table:results}). We also find that on a observation-level, the allocation is an accurate estimator of sufficiency category, with a cross-dataset average of $76\%$ and with few datasets with accuracy under $60\%$.

\begin{table*}[t] 
\centering
\small
\begin{tabular}{ |p{2.2cm}|p{1.25cm}|p{1.25cm}|p{1.25cm}|p{1.25cm}|p{1.25cm}|p{1.25cm}|p{1.25cm}|p{1.25cm}|p{1.25cm}| }
 \hline
 Dataset & AUC & PPCR &  PQEOM & PQOM & PCFA & 95TQM & Max Acc & Argmax $q$ & $s$ Acc\\
 \hline
  Wine & 79 $\pm$ 0 & 21 $\pm$ 0 & 71 $\pm$ 0 & 0 $\pm$ 0 & 7 $\pm$ 0 & 98 $\pm$ 0 & 80 $\pm$ 0 & 70 $\pm$ 0 & 78 $\pm$ 0 \\
 \hline
  Phoneme & 87 $\pm$ 0 & 12 $\pm$ 1 & 78 $\pm$ 4 & 7 $\pm$ 2 & 6 $\pm$ 1 & 100 $\pm$ 0 & 87 $\pm$ 0 & 50 $\pm$ 34 & 81 $\pm$ 2 \\
 \hline
  KDDIPUMS & 88 $\pm$ 0 & 17 $\pm$ 1 & 65 $\pm$ 3 & 34 $\pm$ 5 & 17 $\pm$ 7 & 100 $\pm$ 0 & 88 $\pm$ 0 & 66 $\pm$ 9 & 80 $\pm$ 1 \\
 \hline
  EyeMovements & 66 $\pm$ 0 & 33 $\pm$ 0 & 55 $\pm$ 1 & 7 $\pm$ 6 & 15 $\pm$ 2 & 70 $\pm$ 0 & 68 $\pm$ 0 & 31 $\pm$ 24 & 52 $\pm$ 0 \\
 \hline
  Pol & 98 $\pm$ 0 & 49 $\pm$ 0 & 98 $\pm$ 0 & 2 $\pm$ 0 & 0 $\pm$ 0 & 100 $\pm$ 0 & 99 $\pm$ 0 & 98 $\pm$ 0 & 96 $\pm$ 0 \\
 \hline
  Bank & 76 $\pm$ 0 & -19 $\pm$ 0 & 4 $\pm$ 1 & 0 $\pm$ 0 & 0 $\pm$ 0 & 100 $\pm$ 0 & 79 $\pm$ 0 & 100 $\pm$ 0 & 71 $\pm$ 0 \\
 \hline
  MagicTelescope & 86 $\pm$ 0 & 39 $\pm$ 0 & 87 $\pm$ 2 & 12 $\pm$ 11 & 10 $\pm$ 0 & 100 $\pm$ 0 & 86 $\pm$ 0 & 47 $\pm$ 28 & 82 $\pm$ 1 \\
 \hline
  House16H & 89 $\pm$ 0 & 40 $\pm$ 0 & 84 $\pm$ 4 & 9 $\pm$ 6 & 6 $\pm$ 2 & 98 $\pm$ 0 & 89 $\pm$ 0 & 82 $\pm$ 8 & 86 $\pm$ 0 \\
 \hline
  Credit & 78 $\pm$ 0 & 5 $\pm$ 1 & 56 $\pm$ 4 & 14 $\pm$ 19 & 95 $\pm$ 0 & 100 $\pm$ 0 & 78 $\pm$ 0 & 76 $\pm$ 6 & 72 $\pm$ 0 \\
 \hline
  California & 90 $\pm$ 0 & 52 $\pm$ 0 & 88 $\pm$ 0 & 0 $\pm$ 0 & 7 $\pm$ 0 & 98 $\pm$ 0 & 91 $\pm$ 0 & 88 $\pm$ 0 & 90 $\pm$ 0 \\
 \hline
  Electricity & 92 $\pm$ 0 & 58 $\pm$ 0 & 88 $\pm$ 0 & 0 $\pm$ 0 & 7 $\pm$ 0 & 98 $\pm$ 0 & 93 $\pm$ 0 & 88 $\pm$ 0 & 92 $\pm$ 0 \\
 \hline
  Jannis & 79 $\pm$ 0 & 30 $\pm$ 0 & 53 $\pm$ 5 & 21 $\pm$ 5 & 14 $\pm$ 1 & 98 $\pm$ 0 & 79 $\pm$ 0 & 35 $\pm$ 6 & 76 $\pm$ 0 \\
 \hline
  MiniBooNE & 94 $\pm$ 0 & 54 $\pm$ 0 & 90 $\pm$ 0 & 0 $\pm$ 0 & 5 $\pm$ 0 & 100 $\pm$ 0 & 94 $\pm$ 0 & 90 $\pm$ 0 & 92 $\pm$ 0 \\
 \hline
  WineR & 73 $\pm$ 0 & 43 $\pm$ 0 & 85 $\pm$ 0 & 10 $\pm$ 0 & 20 $\pm$ 0 & 90 $\pm$ 0 & 74 $\pm$ 0 & 78 $\pm$ 0 & 67 $\pm$ 0 \\
 \hline
  IsoletR & 91 $\pm$ 0 & 89 $\pm$ 0 & 68 $\pm$ 0 & 0 $\pm$ 0 & 49 $\pm$ 10 & 73 $\pm$ 0 & 95 $\pm$ 0 & 68 $\pm$ 0 & 87 $\pm$ 1 \\
 \hline
  CPUR & 75 $\pm$ 0 & 40 $\pm$ 2 & 52 $\pm$ 18 & 0 $\pm$ 0 & 29 $\pm$ 11 & 80 $\pm$ 0 & 77 $\pm$ 0 & 70 $\pm$ 0 & 59 $\pm$ 1 \\
 \hline
  SulfurR & 98 $\pm$ 0 & 63 $\pm$ 2 & 73 $\pm$ 9 & 1 $\pm$ 2 & 79 $\pm$ 4 & 100 $\pm$ 0 & 98 $\pm$ 0 & 84 $\pm$ 22 & 96 $\pm$ 0 \\
 \hline
  BrazilianHousesR & 96 $\pm$ 0 & 71 $\pm$ 1 & 83 $\pm$ 0 & 64 $\pm$ 6 & 1 $\pm$ 2 & 93 $\pm$ 0 & 98 $\pm$ 0 & 8 $\pm$ 3 & 88 $\pm$ 0 \\
 \hline
  AileronsR & 75 $\pm$ 0 & 13 $\pm$ 0 & 70 $\pm$ 10 & 5 $\pm$ 7 & 4 $\pm$ 2 & 100 $\pm$ 0 & 76 $\pm$ 0 & 40 $\pm$ 35 & 64 $\pm$ 0 \\
 \hline
  MiamiHousingR & 76 $\pm$ 0 & 44 $\pm$ 0 & 76 $\pm$ 0 & 0 $\pm$ 0 & 67 $\pm$ 10 & 80 $\pm$ 0 & 78 $\pm$ 0 & 75 $\pm$ 0 & 65 $\pm$ 0 \\
 \hline
  PolR & 88 $\pm$ 0 & 44 $\pm$ 1 & 96 $\pm$ 1 & 93 $\pm$ 1 & 88 $\pm$ 0 & 100 $\pm$ 0 & 88 $\pm$ 0 & 81 $\pm$ 2 & 84 $\pm$ 0 \\
 \hline
  ElevatorsR & 75 $\pm$ 0 & 25 $\pm$ 0 & 64 $\pm$ 4 & 2 $\pm$ 2 & 17 $\pm$ 0 & 88 $\pm$ 0 & 75 $\pm$ 0 & 51 $\pm$ 29 & 63 $\pm$ 0 \\
 \hline
  BikeSharingR & 77 $\pm$ 0 & 26 $\pm$ 1 & 88 $\pm$ 4 & 22 $\pm$ 16 & 82 $\pm$ 1 & 100 $\pm$ 0 & 78 $\pm$ 0 & 21 $\pm$ 13 & 72 $\pm$ 0 \\
 \hline
  FifaR & 77 $\pm$ 0 & 33 $\pm$ 0 & 95 $\pm$ 0 & 93 $\pm$ 0 & 78 $\pm$ 0 & 100 $\pm$ 0 & 77 $\pm$ 0 & 50 $\pm$ 0 & 69 $\pm$ 0 \\
 \hline
  CaliforniaR & 78 $\pm$ 0 & 38 $\pm$ 0 & 73 $\pm$ 0 & 68 $\pm$ 0 & 68 $\pm$ 0 & 93 $\pm$ 0 & 79 $\pm$ 0 & 42 $\pm$ 0 & 63 $\pm$ 0 \\
 \hline
  HousesR & 78 $\pm$ 0 & 41 $\pm$ 0 & 73 $\pm$ 0 & 0 $\pm$ 0 & 48 $\pm$ 1 & 80 $\pm$ 0 & 79 $\pm$ 0 & 73 $\pm$ 0 & 62 $\pm$ 0 \\
 \hline
  SuperconductR & 83 $\pm$ 0 & 42 $\pm$ 0 & 60 $\pm$ 13 & 24 $\pm$ 11 & 95 $\pm$ 0 & 95 $\pm$ 0 & 83 $\pm$ 0 & 57 $\pm$ 1 & 76 $\pm$ 0 \\
 \hline
  HouseSalesR & 76 $\pm$ 0 & 50 $\pm$ 1 & 64 $\pm$ 12 & 0 $\pm$ 0 & 56 $\pm$ 6 & 85 $\pm$ 0 & 78 $\pm$ 0 & 78 $\pm$ 0 & 63 $\pm$ 0 \\
 \hline
  House16HR & 92 $\pm$ 0 & 55 $\pm$ 0 & 90 $\pm$ 0 & 0 $\pm$ 0 & 6 $\pm$ 0 & 98 $\pm$ 0 & 92 $\pm$ 0 & 90 $\pm$ 0 & 86 $\pm$ 0 \\
 \hline
  DiamondsR & 70 $\pm$ 0 & 19 $\pm$ 0 & 83 $\pm$ 0 & 34 $\pm$ 6 & 73 $\pm$ 0 & 100 $\pm$ 0 & 71 $\pm$ 0 & 20 $\pm$ 6 & 65 $\pm$ 0 \\
 \hline
  MedicalChargesR & 86 $\pm$ 0 & 24 $\pm$ 0 & 91 $\pm$ 1 & 85 $\pm$ 1 & 0 $\pm$ 0 & 100 $\pm$ 0 & 86 $\pm$ 0 & 67 $\pm$ 2 & 83 $\pm$ 0 \\
 \hline
  \textbf{Average} & \textbf{83 $\pm$ 9} & \textbf{37 $\pm$ 21} & \textbf{74 $\pm$ 19} & \textbf{20 $\pm$ 29} & \textbf{35 $\pm$ 34} & \textbf{94 $\pm$ 9} & \textbf{84 $\pm$ 8} & \textbf{64 $\pm$ 24} & \textbf{76 $\pm$ 12} \\
 \hline
\end{tabular}
\caption{This table summarizes allocation performance across several metrics on each dataset and, in the bottom row, across the datasets (all metrics are reported as percentages). Averages and standard deviations are reported over 5 replicates. Metric definitions can be found in the Metrics section, and discussion of results can be found in the Results section.}

\label{table:results}
\end{table*}

\subsection{Ablation Studies}

\subsubsection{Allocator Feature Set Selection}
In addition to the features $x$ used to learn the glass box and black box models, the allocation task also has access to their predictions $g(x)$ and $b(x)$, and any functions of the two - since the allocator is learned subsequent to the training of these models. To obtain the optimal feature set for allocation, standard tuning procedures (e.g. cross validation) can be employed to evaluate all feature sets of interest. However, as each candidate feature set requires the training of a corresponding allocator for evaluation, this approach can be prohibitively costly. 

Thus, the following study was conducted to determine whether a consistently best feature set exists for the tabular data context used in the experiments. First, the universe of candidate features was selected to be the original features $x$ used as inputs for the ensemble component models, the predictions of both of these models $g(x)$ and $b(x)$, and finally two measures of discrepancy between the predictions, the cross-entropy $d_{ce}(g(x),b(x))$ and the mean squared error $d_{mse}(g(x),b(x))$. The measures of disagreement were included as features as they translate to the ``feature independent" strategy of allocation to model $b$ for low values of $d(a(x),b(x))$ - in other words the optimal allocation strategy assuming $a$ is always correct.

The candidate features were grouped into the sets listed in Table \ref{table:feature_set_selection} and then used to train allocators on a subset of the benchmark datasets  (Wine, WineR, Phoneme, SulfurR, Bank, BrazilianHousesR, FifaR, KDDIPUMS) with 6 replicates per model.

Next each feature set was evaluated as follows. First, within each dataset, each feature set's performance (defined as the AUC) was compared to the performance of the best alternative set of features. Then, the proportion of datasets for which the feature set being evaluated was not significantly worse (i.e. either significantly better or not significantly different) than the best alternative was recorded and reported in Table \ref{table:feature_set_selection} for three significance levels ($10\%$, $5\%$, $1\%$).

The results support the following three conclusions. First, no one feature set proved universally best across the tested datasets and thus a full search across feature sets would be advised in settings without resource constraint. Second, although no universally best feature set was found, the ``kitchen sink" set of all candidate features ($x$, $g$, $b$, $d_{ce}$, $d_{mse}$) was found to be best most consistently and was thus used to train all allocators reported in Table \ref{table:results}. Finally, allocators trained on just the original features $x$ were found to be consistently worst among all alternatives thus supporting the augmentation of the original features in some form. This finding is consistent with the intuition that the predictions of the component models would be very useful to learning the optimal allocation and would be either very difficult or impossible to learn from $r^{}$, the optimal allocation ranking response, alone during training. 

\begin{table}
    \centering
    \begin{tabular}{|p{2.33cm}|c|c|c|}
         \hline
         Feature Set & $\alpha:0.01$ & $\alpha:0.05$ & $\alpha:0.1$ \\
         \hline
         $x$ & 18.75\% & 18.75\% & 12.50\%  \\
         \hline
         $g$, $b$ & 43.75\% & 43.75\% & 31.25\%  \\
         \hline
         $d_{ce}$ & 31.25\% & 18.75\% & 18.75\%  \\
         \hline
         $d_{mse}$ & 50.00\% & 43.75\% & 37.50\%  \\
         \hline
         $x$, $d_{ce}$ & 37.50\% & 25.00\% & 18.75\%  \\
         \hline
         $x$, $d_{mse}$ & 56.25\% & 37.50\% & 25.00\%  \\
         \hline
         $g$, $b$, $d_{ce}$ & 31.25\% & 25.00\% & 25.00\%  \\
         \hline
         $g$, $b$, $d_{mse}$ & 50.00\% & 50.00\% & 37.50\%  \\
         \hline
         $x$, $g$, $b$ & 56.25\% & 43.75\% & 37.50\%  \\
         \hline
         $x$, $g$, $b$, $d_{ce}$ & 50.00\% & 43.75\% & 37.50\%  \\
         \hline
         $x$, $g$, $b$, $d_{mse}$ & 56.25\% & 37.50\% & 37.50\%  \\
         \hline
         $x$, $g$, $b$, $d_{ce}$, $d_{mse}$ & \textbf{75.00\%} &  \textbf{56.25\%} & \textbf{43.75\%}  \\
         \hline
    \end{tabular}
    \caption{This table reports the percentage of datasets for which the allocator learned on the corresponding feature set is significantly better than or not significantly different from the best alternative feature set trained allocator. Results across three significance levels are reported and show that the ``kitchen sink" $x,g,b,d_{ce},d_{mse}$ feature set is most consistently best (bolded) while the ``unaugmented" original feature set of $x$ is consistently the worst across all $\alpha$.}
    \label{table:feature_set_selection}
\end{table}

\subsubsection{Ensemble Component Model Selection}

The performance of any allocated ensemble is highly dependent not only on the individual performance of its component models (i.e. $g$ and $b$) but on their level of synergy as well. In particular, it may be the case that the component model pair in $(g_0,b_0,a_0)$ individually outperforms the respective component models in $(g_1,b_1,a_1)$ but that the allocator $a_0$ trained with $(g_0,b_0)$ underperforms $a_1$. In this case, the high relative advantages of $(g_1,b_1)$ in different segments of the feature space overcome their global performance disadvantages as individual models compared to their counterparts in $(g_0,b_0)$ to yield a stronger ensemble.

Thus, to determine how often high relative advantage overcomes superior individual performance in allocator training, the following study was conducted. For each dataset, an allocator was trained on each combination of available glass box (tree and regression) and black box (gradient boosting trees and neural network) models (i.e. four allocators per dataset). Then the allocator $a_I$, trained using the pair of component models $(g_I,b_I)$ with the highest individual validation performance, was identified along with the allocator $a_C$, trained using the pair of component models $(g_C,b_C)$ resulting in the highest ensemble validation performance. Finally the difference in test performance was measured between $a_C$ and $a_I$ ($AUC \Delta = AUC(a_C)-AUC(a_I)$). 

The resulting $AUC \Delta$ values support the following two conclusions. Firstly, while a relatively high proportion ($41.9\%$) of datasets yield different allocators depending on which of the two different component model selection processes (individual vs. combined performance) they utilize, the cross-dataset average difference in allocator performance is not significantly different from zero ($0.01 \pm 0.03$). This result suggests that the glass box and black box model types used for the experiments did not exhibit high relative expertise in different parts of the feature space, indicating that it may be beneficial to use a more diverse set of component models in this setting. However, in rare cases (e.g. IsoletR, BrazilianHousesR) the combined performance selection method yields as much as $15\%$ in additional performance. Thus, in resource constrained settings, or in cases in which many glass box and black box model types are under consideration, the individual performance selection method appears relatively low risk, although a full search across all component model combinations (the method used for Table \ref{table:results}) is recommended when feasible.

\section{Acknowledgements}
We would like to thank Jonathan Siegel for valuable discussion of the theoretical aspects of this work.

This research is supported by the National Science Foundation under grant number CCF-2205004. Computations for this research were performed on the Pennsylvania State University’s Institute for Computational and Data Sciences’ Roar supercomputer.


\bibliography{aaai24}

\onecolumn
\appendix


\begingroup 
\linespread{1.6}\selectfont 

\section{Appendix}

\subsection{A. Theoretical Results}

\setcounter{proposition}{0}

\begin{definition}{(Z Sets)}\\
$Z^{}_0 = \{z:z\in \mathcal{Z}, s^{}_g(z)+s^{}_b(z)=0 \}$\\
$Z^{}_2 = \{z:z\in \mathcal{Z}, s^{}_g(z)+s^{}_b(z)=2 \}$\\ 
$Z^{}_g = \{z:z\in \mathcal{Z}, s^{}_g(z)=1,s^{}_b(z)=0 \}$\\ 
$Z^{}_b = \{z:z\in \mathcal{Z}, s^{}_g(z)=0,s^{}_b(z)=1 \}$\\
$Z_{\bigtriangleup,\square} = Z_{\bigtriangleup} \times Z_{\square}$
\end{definition}

\begin{definition}{(Sufficiency)}\\
$S(a) = \frac{1}{n}\sum^{n}_{i=1}s(a,v_i)$\\
$s(a,v) = s_g(z)a(v)+s_b(z)(1-a(v))$\\
$S_{g}(a) = \frac{1}{n}\sum^{n}_{i=1}s_g(z_i)a(v_i)$
\end{definition}

\begin{lemma}{($Z$ Sets Ordered By Ranking)}\label{lemma_incr_r_tilde_appendix}\\
$\forall (z_{b},z_{0},z_{2},z_{g}) \in (Z_{b} \times Z_{0} \times Z_{2} \times Z_{g}), r(z_b)<r(z_0)<r(z_2)<r(z_g)$
\end{lemma}

\begin{proof}
Notice $\{r(z_b) \colon z_b \in Z_b\} \subseteq (-2,-1], 
\{r(z_0) \colon z_0 \in Z_0\} \subseteq (-1,0], 
\{r(z_2) \colon z_2 \in Z_2\} \subseteq (0,1], 
\{r(z_g) \colon z_g \in Z_g\} \subseteq (1,2]$, statement follows.
\end{proof}

\begin{lemma}{(Allocated By Top Ranked)}\label{lemma_Za_is_Rtilde_appendix}\\
$ |V^{\prime}_{q}| = n_q, V^{\prime}_q = R_q$ where $V^{\prime}_q = \{v \colon a^{\prime}_{q}(v)=1\}, R_{q} = \{z\colon \mathrm{rank}(r(z))>n_{1-q}\}$
\end{lemma}

\begin{proof}
\underline{Part 1}: $V^{\prime}_{q} = \{v \colon a^{\prime}_{q}(v)=1\}=\{z\colon \tilde{r}(z)>1-q\}=\{z \colon \frac{\mathrm{rank}(r(z))}{n}>1-q\}=\{z:\mathrm{rank}(r(z))>n(1-q)\}=\{z:\mathrm{rank}(r(z))>n_{1-q}\}=R_{q}$ \\
\underline{Part 2}: Observe that, by construction, $\tilde{r}(z)$ is the $\tilde{r}(z)^{th}$ percentile. Thus, $|V^{\prime}_{q}|=|\{z\colon \tilde{r}(z)>1-q\}|=n(1-(1-q))=n_{q}$.
\end{proof}

\begin{proposition}{(Maximal Sufficient Performance)}\label{thm_1_perfmax_appendix}\\
$\forall q \in \mathcal{Q}, a^{\prime}_{q} \in A^{*}_{q}$ where $A^{*}_{q}=\{a^{*}_{q} \colon a^{*}_{q}=\arg\max_{a_{q}\in A_{q}} S(a_q)\}$
\end{proposition}

\begin{proof} Assume for contradiction that $\exists b_q \in A^*_q$ s.t. $S(b_q)>S(a^{\prime}_q)$.\\
$\implies \exists v_i,v_j \in V^n$ s.t. $s(b_q,v_i)+s(b_q,v_j)>s(a^{\prime}_q,v_i)+s(a^{\prime},v_j), b_q(v_i) \neq a^{\prime}_q(v_i), b_q(v_j) \neq a^{\prime}_q(v_j)$\\
w.l.o.g. take $a^{\prime}_q(v_i)>a^{\prime}_q(v_j)$ and $b_q(v_i)<b_q(v_j)$\\
then $s(b_q,v_i)+s(b_q,v_j)=s_b(z_i)+s_g(z_j)$ and $s(a^{\prime}_q,v_i)+s(a^{\prime},v_j)=s_g(z_i)+s_b(z_j)$\\
Now, taking it by case:\\
for $(z_i,z_j) \in Z_k \times Z_k, k \in \{0,2,g,b\}, s_b(z_i)+s_g(z_j) = s_g(z_i)+s_b(z_j) \Rightarrow\Leftarrow$\\
for $(z_i,z_j) \in Z_{g,0} \cup Z_{0,b} \cup Z_{g,b} \cup Z_{g,2} \cup Z_{2,b}, s_b(z_i)+s_g(z_j) < s_g(z_i)+s_b(z_j) \Rightarrow\Leftarrow$\\
for $(z_i,z_j) \in Z_{0,g} \cup Z_{b,0} \cup Z_{b,g} \cup Z_{2,g} \cup Z_{b,2}, r(z_i)<r(z_j) \implies a^{\prime}_q(z_i) \leq a^{\prime}_q(z_j) \Rightarrow\Leftarrow$\\
$\implies \nexists b_q \in A^*_q$ s.t. $S(b_q)>S(a^{\prime}_q) \implies a^{\prime}_q \in A^*_q$
\end{proof}

\newpage

\begin{proposition}{(Maximal Sufficient Explainable Performance)}\label{thm_2_perfmaxg_appendix}\\
$\forall q \in \mathcal{Q}, a^{\prime}_{q} \in A^{*}_{q\vert g}$ where $A^{*}_{q\vert g}=\{a^{*}_{q\vert g} \colon a^{*}_{q\vert g}=\arg\max_{a_{q}\in A_{q}} S_{g}(a_q)\}$
\end{proposition}

\begin{proof}
Assume for contradiction that $\exists b_q \in A^*_{q\vert g}$ s.t. $S_g(b_q)>S_g(a^{\prime}_q)$\\
$\implies \exists v_i,v_j \in V^n$ s.t. $s_g(z_i)b_q(v_i)+s_g(z_j)b_q(v_j)>s_g(z_i)a^{\prime}_q(v_i)+s_g(z_j)a^{\prime}_q(v_j), b_q(v_i) \neq a^{\prime}_q(v_i), b_q(v_j) \neq a^{\prime}_q(v_j)$\\
w.l.o.g. take $a^{\prime}_q(v_i)>a^{\prime}_q(v_j)$ and $b_q(v_i)<b_q(v_j)$\\
then $s_g(z_i)b_q(v_i)+s_g(z_j)b_q(v_j)=s_g(z_j)$ and $s_g(z_i)a^{\prime}_q(v_i)+s_g(z_j)a^{\prime}_q(v_j)=s_g(z_i)$\\
Now, taking it by case: \\
for $(z_i,z_j) \in Z_{k,k},k\in \{0,2,g,b\}, s_g(z_i)=s_g(z_j) \Rightarrow\Leftarrow$\\
for $(z_i,z_j) \in Z_{b,0}\cup Z_{2,g} \cup Z_{0,b}\cup Z_{g,2}, s_g(z_i)=s_g(z_j) \Rightarrow\Leftarrow$\\
for $(z_i,z_j) \in Z_{g,0}\cup Z_{2,0} \cup Z_{g,b}\cup Z_{2,b}, s_g(z_i)>s_g(z_j) \Rightarrow\Leftarrow$\\
for $(z_i,z_j) \in Z_{0,g}\cup Z_{0,2} \cup Z_{b,g}\cup Z_{b,2}, r(z_i)<r(z_j) \implies a^{\prime}_q(v_i)\leq a^{\prime}_q(v_j) \Rightarrow\Leftarrow$\\
$\implies \nexists b_q\in A^*_{q|g}$ s.t. $S_g(b_q)>S_g(a^{\prime}_q) \implies a^{\prime}_q \in A^*_{q|g}$
\end{proof}

\begin{proposition}{(Conditional Maximum Absolute Performance)}\label{thm_3_absperf_appendix}\\
$\forall q \in \mathcal{Q}, a^{\prime}_{q} \in A^{*}_{q\vert s}$ where $A^{*}_{q\vert s}=\{a^{*}_{q\vert s} \colon a^{*}_{q\vert s}=\arg\min_{a_{q}\in A^*_{q}\cap A^*_{q\vert g}} L(a_q)\}$,\\ $L(a_q) = \frac{1}{n}\sum l(g(x_i),y_i)a_q(v_i)+l(b(x_i),y_i)(1-a_q(v_i))$
\end{proposition}

\begin{proof}
Assume for contradiction that $\exists b_q \in A^{*}_{q\vert s}$ s.t. $L(b_q)<L(a^{\prime}_q)$\\
$\implies \exists v_i,v_j \in V^n$ s.t. $l_i(a^{\prime}_q)+l_j(a^{\prime}_q) > l_i(b_q)+l_j(b_q)$\\
Now, taking $(v_i,v_j)$ by case:\\
for $(v_i,v_j) \in Z_{m,n}, m\neq n$, we have $a^{\prime}_q(v_i)=b_q(v_i),a^{\prime}_q(v_j)=b_q(v_j)$, as both $a^{\prime}_q, b_q \in A^*_q\cap A^*_{q\vert g}$\\
$\implies l_i(a^{\prime}_q)+l_j(a^{\prime}_q)=l_i(b_q)+l_j(b_q) \Rightarrow\Leftarrow$\\
for $(v_i,v_j) \in Z_{k,k}$, w.l.o.g. take $a^{\prime}_q(v_i)>a^{\prime}_q(v_j)$ and $b_q(v_i)<b_q(v_j)$\\
then $l_i(a^{\prime}_q)+l_j(a^{\prime}_q) > l_i(b_q)+l_j(b_q) \Longleftrightarrow l(g(x_i),y_i)+l(b(x_j),y_j) > l(b(x_i),y_i)+l(g(x_j),y_j)$\\ 
however, $a^{\prime}_q(v_i)>a^{\prime}_q(v_j) \implies r(z_i)>r(z_j)$\\
$\Longleftrightarrow -\sigma(l(g(x_i),y_i)-l(b(x_i),y_i))>-\sigma(l(g(x_j),y_j)-l(b(x_j),y_j))$\\
$\Longleftrightarrow l(g(x_i),y_i)+l(b(x_j),y_j)<l(b(x_i),y_i)+l(g(x_j),y_j) \Rightarrow\Leftarrow$\\
$\implies \nexists b_q \in A^*_{q\vert s}$ s.t. $L(b_q)<L(a^{\prime}_q) \implies a^{\prime}_q \in A^*_{q\vert s}$
\end{proof}

\begin{proposition}{(Monotone Allocation)}\label{thm_4_monotone_appendix}\\
$\forall q_i<q_j \in \mathcal{Q}$, $\{v:v\in\mathcal{V},a^{\prime}_{q_i}(v)=1\}\subseteq \{v:v\in\mathcal{V},a^{\prime}_{q_j}(v)=1\}$
\end{proposition}

\begin{proof}
From Lemma \ref{lemma_Za_is_Rtilde_appendix}, $\{v:v\in\mathcal{V},a^{\prime}_{q_i}(v)=1\} = \{z:z\in\mathcal{Z},\mathrm{rank}_{D^n}(r(z))>n_{1-q_i}\}$. Then since $q_i<q_j, n_{1-q_i}>n_{1-q_j}$, thus $\{z:z\in\mathcal{Z},a^{\prime}_{q_i}(z)=1\} =\{z:z\in\mathcal{Z},\mathrm{rank}_{D^n}(r(z))>n_{1-q_i}\} \subseteq \{z:z\in\mathcal{Z},\mathrm{rank}_{D^n}(r(z))>n_{1-q_j}\} = \{v:v\in\mathcal{V},a^{\prime}_{q_j}(v)=1\}$.
\end{proof}


\newpage

\subsection{B. Dataset Details}
\begingroup
\renewcommand{\arraystretch}{0.8}
\begin{table*}[h] 
\centering
\footnotesize
\begin{tabular}{ |l|l|l|l|l|l|l|l|l| }
 \hline
 Dataset & Task & Observations & Features & Epochs & Tree & Regr. & GBT & DNN \\
 \hline
  Wine & classification & 2,554 & 11 & 800 & 73.9 & 72.1 & \textbf{80.5} & 79.5 \\
 \hline
  Phoneme & classification & 3,172 & 5 & 800 & 85.3 & 73.3 & \textbf{88.5} & 87.1 \\
 \hline
  KDDIPUMS & classification & 5,188 & 20 & 800 & \textbf{88.3} & 85.8 & 88.0 & 83.6 \\
 \hline
  EyeMovements & classification & 7,608 & 20 & 800 & 58.5 & 53.7 & \textbf{68.2} & 56.6 \\
 \hline
  Pol & classification & 10,082 & 26 & 400 & 97.1 & 86.9 & 98.4 & \textbf{98.5} \\
 \hline
  Bank & classification & 10,578 & 7 & 400 & 79.0 & 73.6 & \textbf{79.7} & 75.6 \\
 \hline
  MagicTelescope & classification & 13,376 & 10 & 400 & 81.7 & 77.6 & \textbf{85.9} & 84.4 \\
 \hline
  House16H & classification & 13,488 & 16 & 400 & 84.7 & 83.2 & \textbf{89.2} & 86.2 \\
 \hline
  Credit & classification & 16,714 & 10 & 400 & 77.1 & 73.9 & \textbf{77.7} & 71.4 \\
 \hline
  California & classification & 20,634 & 8 & 400 & 85.5 & 82.4 & \textbf{90.6} & 87.8 \\
 \hline
  Electricity & classification & 38,474 & 7 & 400 & 86.8 & 74.6 & \textbf{92.6} & 81.5 \\
 \hline
  Jannis & classification & 57,580 & 54 & 400 & 74.6 & 72.4 & \textbf{79.3} & 77.6 \\
 \hline
  MiniBooNE & classification & 72,998 & 50 & 400 & 89.8 & 84.6 & \textbf{93.9} & 93.0 \\
 \hline
  WineR & regression & 6,497 & 11 & 800 & 0.25 & 0.25 & \textbf{0.20} & 0.23 \\
 \hline
  IsoletR & regression & 7,797 & 613 & 800 & 0.39 & 0.38 & 0.25 & \textbf{0.16} \\
 \hline
  CPUR & regression & 8,192 & 21 & 800 & 0.06 & 0.19 & \textbf{0.04} & 0.05 \\
 \hline
  SulfurR & regression & 10,081 & 6 & 400 & 0.06 & 0.08 & 0.05 & \textbf{0.03} \\
 \hline
  BrazilianHousesR & regression & 10,692 & 8 & 400 & 0.02 & 0.09 & 0.01 & \textbf{0.01} \\
 \hline
  AileronsR & regression & 13,750 & 33 & 400 & 0.11 & 0.10 & \textbf{0.09} & 0.09 \\
 \hline
  MiamiHousingR & regression & 13,932 & 14 & 400 & 0.12 & 0.17 & \textbf{0.08} & 0.08 \\
 \hline
  PolR & regression & 15,000 & 26 & 400 & 0.15 & 0.61 & 0.09 & \textbf{0.07} \\
 \hline
  ElevatorsR & regression & 16,599 & 16 & 400 & 0.10 & 0.09 & 0.07 & \textbf{0.06} \\
 \hline
  BikeSharingR & regression & 17,379 & 6 & 400 & 0.22 & 0.30 & \textbf{0.21} & 0.21 \\
 \hline
  FifaR & regression & 18,063 & 5 & 400 & 0.25 & 0.34 & \textbf{0.23} & 0.24 \\
 \hline
  CaliforniaR & regression & 20,640 & 8 & 400 & 0.23 & 0.27 & \textbf{0.16} & 0.18 \\
 \hline
  HousesR & regression & 20,640 & 8 & 400 & 0.17 & 0.20 & \textbf{0.13} & 0.13 \\
 \hline
  SuperconductR & regression & 21,263 & 79 & 400 & 0.14 & 0.21 & \textbf{0.10} & 0.11 \\
 \hline
  HouseSalesR & regression & 21,613 & 15 & 400 & 0.10 & 0.12 & \textbf{0.08} & 0.08 \\
 \hline
  House16HR & regression & 22,784 & 16 & 400 & 0.10 & 0.11 & \textbf{0.08} & 0.10 \\
 \hline
  DiamondsR & regression & 53,940 & 6 & 400 & 0.12 & 0.14 & 0.12 & \textbf{0.11} \\
 \hline
  MedicalChargesR & regression & 163,065 & 5 & 100 & 0.04 & 0.12 & 0.04 & \textbf{0.04} \\
 \hline
\end{tabular}
\caption{This table summarizes the datasets used in the experiments - their number of observations and features and the number of training epochs used to learn the neural networks for each. Also included are the test accuracies (for classification tasks) and RMSE (for regression tasks) for each ensemble component model with best performance bolded.} 
\label{table:ds_summary}
\end{table*}
\endgroup


\newpage

\subsection{C. Hyperparameter Tuning}

\begin{table*}[h]
\centering
\normalsize
\begin{tabular}{ |l|l|l|l|l| }
 \hline
 Model & L1 penalty $\beta$ & min split & max leaf & max depth \\
 \hline
  Logistic regression & $2^i$ for $i$ in range(-10,10) & - & - & -  \\
 \hline
 Linear regression & $2^i$ for $i$ in range(-10,10) & - & - & - \\
 \hline
 Classification tree & - & $2^i$ for $i$ in range(1,12) & $2^i$ for $i$ in range(1,12) & $2^i$ for $i$ in range(1,12) \\
 \hline
 Regression tree & - & $2^i$ for $i$ in range(1,12) & $2^i$ for $i$ in range(1,12) & $2^i$ for $i$ in range(1,12) \\
 \hline
\end{tabular}
\caption{This table summarizes the hyperparameter values searched by model type for all glass box models.}
\label{table:appendix_tune_gb}
\end{table*}

\begin{table*}[h]
\centering
\normalsize
\begin{tabular}{ |l|l|l|l|l| }
 \hline
 Model & learning rate & n estimators & max depth & subsample \\
 \hline
 GBT classifier & 0.001, 0.01, 0.1 & $2^i$, $i$ in range(4,10) & $2^i$, $i$ in range(3,7) & (2,4,6,8,10)*0.1\\
 \hline
 GBT regressor & 0.001, 0.01, 0.1 & $2^i$, $i$ in range(4,10) & $2^i$, $i$ in range(3,7) & (2,4,6,8,10)*0.1\\
 \hline
 GBT allocator & 0.01, 0.1 & $2^i$, $i$ in range(2,11,2) & $2^i$, $i$ in range(2,6) & (25,50,75,100)*0.01\\
 \hline
\end{tabular}
\caption{This table summarizes the hyperparameter values searched by model type for all gradient boosting trees models.}
\label{table:appendix_tune_gbt}
\end{table*}

\begin{table*}[h]
\centering
\normalsize
\begin{tabular}{ |l|l|l|l|l|l| }
 \hline
 Model & L2 penalty $\beta$ & learning rate & lr schedule & optimizer \\
 \hline
 WRN regressor & 0, 1e-7, 1e-5, 1e-3 & (1e-5, 1e-4, 1e-3), (1e-4, 1e-3, 1e-2) & constant, cosine & Adam, SGD+Mtm \\
 \hline
 WRN allocator & 0, 1e-7, 1e-5, 1e-3 & (1e-5, 1e-4, 1e-3), (1e-4, 1e-3, 1e-2) & constant, cosine & Adam, SGD+Mtm \\
 \hline
\end{tabular}
\caption{This table summarizes the hyperparameter values searched by model type for all neural network models.}
\label{table:appendix_tune_wrn}
\end{table*}

\newpage

\subsection{D. Allocation Tradeoff Curves}

\begin{figure*}[h]%
    \centering
    \includegraphics[width=16cm]{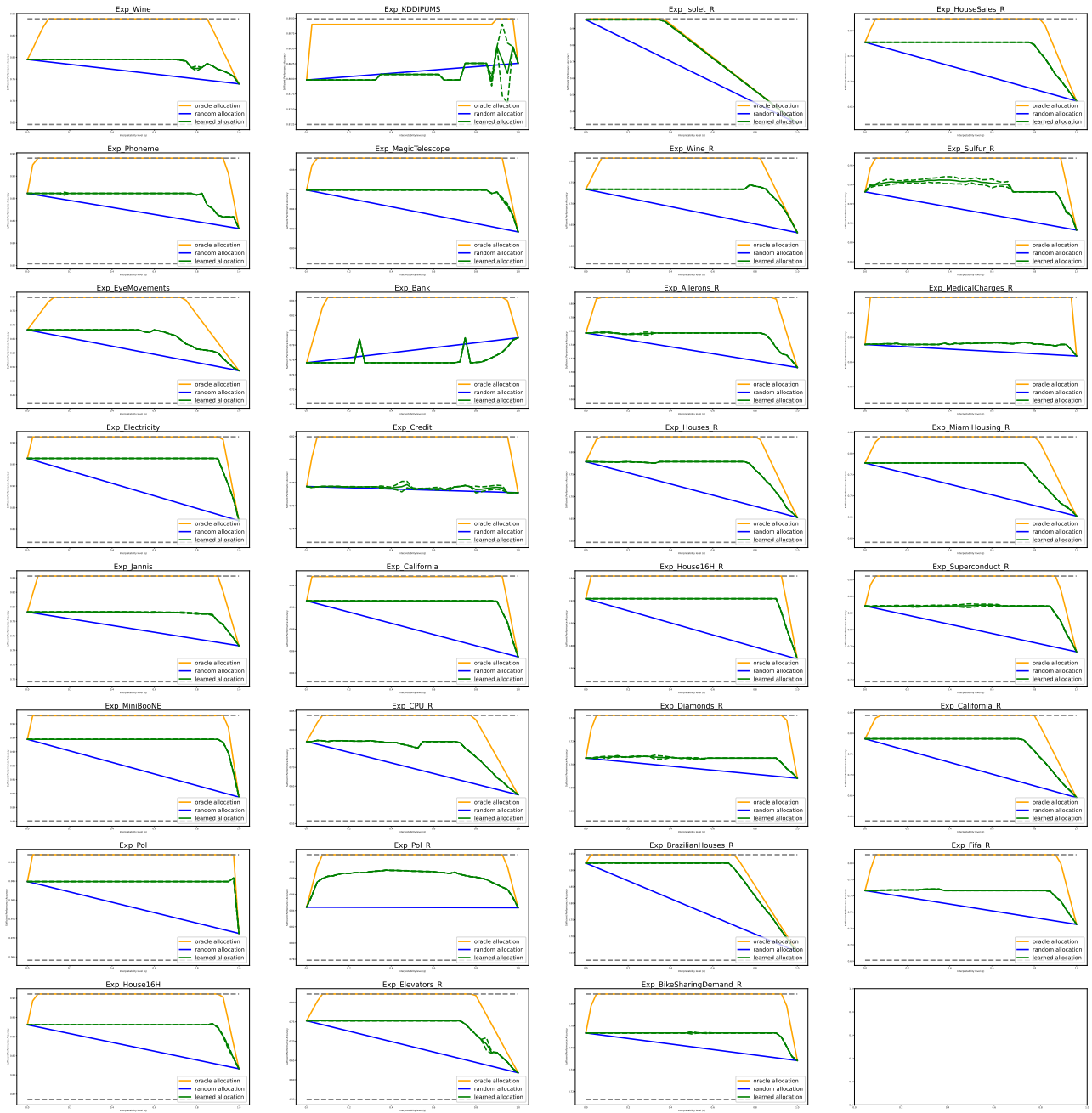}
    \caption{This figure provides the sufficient accuracy values of the allocated EEG ensemble for each explainability level and every dataset.}%
    \label{fig:acc_curves_all}%
\end{figure*}

\endgroup

\end{document}